\documentclass[11pt,a4paper]{article}
\usepackage[hyperref]{acl2021}
\usepackage{times}
\usepackage{latexsym}

\usepackage{microtype}

\aclfinalcopy %

\everypar{\looseness=-1}

\usepackage{caption}
\usepackage{subcaption}
\usepackage{amsmath}
\usepackage{amssymb}
\usepackage{amsfonts}
\usepackage{xfrac}
\usepackage{bbm}
\usepackage{amsthm}
\usepackage{graphicx}
\usepackage{combelow}
\usepackage{qtree}
\usepackage{enumerate}
\usepackage{CJKutf8}
\usepackage{subcaption}
\usepackage{array}

\usepackage{booktabs}
\usepackage{cleveref}
\crefname{section}{\S}{\S\S}
\Crefname{section}{\S}{\S\S}
\crefname{table}{Table}{}
\crefname{figure}{Figure}{}
\crefname{algorithm}{Algorithm}{}
\crefname{equation}{eq.}{eqs.}
\crefname{appendix}{App.}{}
\crefname{prop}{Prop.}{}
\crefformat{section}{\S#2#1#3}  
\newcommand{\citeposs}[1]{\citeauthor{#1}'s (\citeyear{#1})}

\usepackage[disable]{todonotes}

\makeatother

\newcommand{\note}[4][]{\todo[author=#2,color=#3,size=\scriptsize,fancyline,caption={},#1]{#4}} %

\newcommand{\jennifer}[2][]{\note[#1]{jennifer}{blue!40}{#2}}

\newcommand{\calN}{\mathcal{N}}
\newcommand{\calR}{\mathcal{R}}

\newcommand{\VP}{\mathrm{VP}}
\newcommand{\Ss}{\mathrm{S}}
\newcommand{\NP}{\mathrm{NP}}
\newcommand{\PP}{\mathrm{PP}}
\newcommand{\Adj}{\mathrm{Adj}}
\newcommand{\Comp}{\mathrm{Comp}}
\newcommand{\Rel}{\mathrm{Rel}}
\newcommand{\Prep}{\mathrm{Prep}}
\newcommand{\Obj}{\mathrm{Obj}}
\newcommand{\Subj}{\mathrm{Subj}}

\newcommand{\valpha}{\boldsymbol \alpha}

\newcommand{\vb}{\mathbf{b}}

\usepackage{url}

\title{Examining the Inductive Bias of Neural Language Models with Artificial Languages}

\usepackage{emoji}
\newcommand{\ucambridge}{\text{\emoji[twitter]{robot}}}
\newcommand{\ethz}{\text{\emoji[twitter]{speak}}}

\author{
Jennifer C. White\raise1.0ex\hbox{\normalfont\ucambridge}~\;~ 
Ryan Cotterell\raise1.0ex\hbox{\normalfont\ucambridge,\ethz}\\
  \raise1.0ex\hbox{\normalfont\ucambridge}University of Cambridge,~\;~ \raise1.0ex\hbox{\normalfont\ethz}ETH Z\"{u}rich \\
  \texttt{jw2088@cam.ac.uk},~\;~ \texttt{ryan.cotterell@inf.ethz.ch}
}

\date{}

\begin{document}
\maketitle
\begin{abstract}
Since language models are used to model a wide variety of languages, it is natural to ask whether the neural architectures used for the task have inductive biases towards modeling particular types of languages.
Investigation of these biases has proved complicated due to %
the many variables that appear in the experimental setup. %
Languages vary in many typological dimensions, and it is difficult to single out one or two to investigate without the others acting as confounders.
We propose a novel method for investigating the inductive biases of language models using artificial languages.
These languages are constructed to allow us to create parallel corpora across languages that differ only in the typological feature being investigated, such as word order.
We then use them to train and test language models.
This constitutes a fully controlled causal framework, and demonstrates how grammar engineering can serve as a useful tool for analyzing neural models.
Using this method, we find that commonly used neural architectures exhibit different inductive biases: LSTMs display little preference with respect to word ordering, while transformers display a clear preference for some orderings over others.
Further, we find that neither the inductive bias of the LSTM nor that of the transformer appears to reflect any tendencies that we see in attested natural languages.

\end{abstract}

\section{Introduction}
Modern neural architectures used for language modeling,
e.g. Transformer-based language models \citep{vaswani2017attention} and language models based on long-short term memory (LSTM) \citep{hochreiter_lstm, sundermeyer2012lstm},
are intrinsically black boxes. 
This makes it difficult to understand whether their structure leads to an inductive bias which results in certain types of language being easier to learn and model.
To make this point more plainly, we cannot easily conclude much about whether an LSTM language model will perform better on SVO or SOV languages by simply examining its structure.
Moreover, satisfactorily investigating the inductive bias of neural models has the potential to yield useful insight into how they work.
In this work, we explore whether neural language models exhibit biases towards certain types of languages in a novel causal framework through the use of \textbf{artificial languages}. 

\begin{figure}
    \centering
    \includegraphics[scale=0.225]{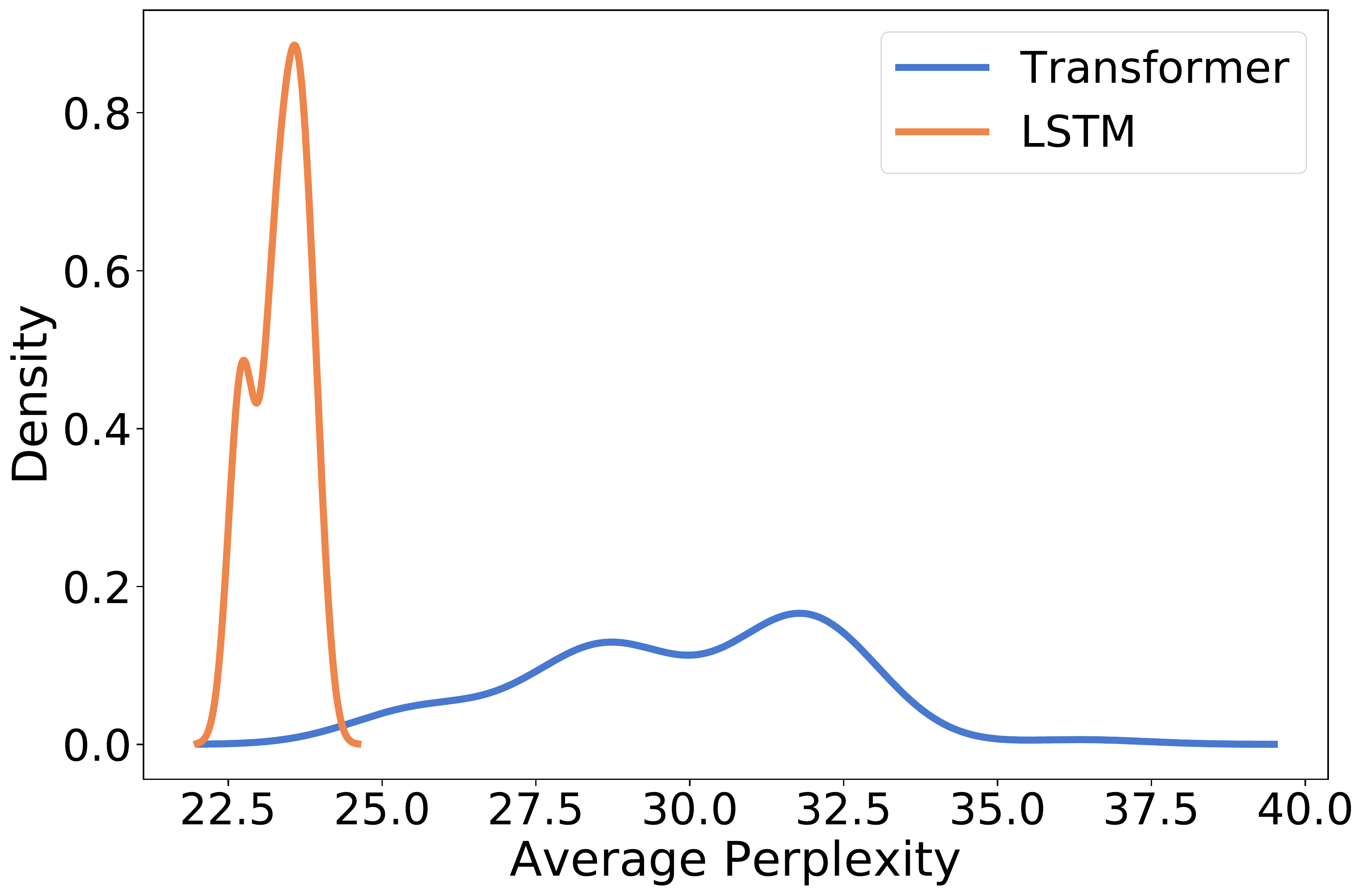}
    \caption{Distribution of average perplexities achieved by transformer- and LSTM-based language models on our artificial languages with varying word order.}
    \label{fig:kde}
\end{figure}
One of the key problems involved in investigating the effect of typological features on language model performance is the difficulty in isolating only the features being investigated, without influence from other features of the languages being investigated or the data being used.
For example, if one were to compare language model performance on English, an SVO language, and Japanese, an SOV language, it would be difficult to directly attribute differences in performance to the difference in word ordering alone.
This is because English and Japanese also differ in many other typological dimensions, such as how subjects are marked, the extent of subject--verb agreement and use of postpositions or prepositions, which could contribute to the difference in performance.
Indeed, recent correlational 
studies have failed to find an effect
between language model performance 
and typological features \cite{cotterell-etal-2018-languages,mielke-etal-2019-kind}.
Moreover, the sentences used for training and testing may differ in content, style or information density, which could further contribute to differences in performance.

Thus, we offer a study investigating the inductive biases of language models through the construction of artificial languages.
Our approach involves creating small context-free grammars resembling subsets of attested languages, which we then use to train and evaluate language models.
In an approach inspired by \citeposs{chomsky1993lectures} framework of principles and parameters, we imbue our grammars with ``switches'' that indicate how to permute the ordering of the non-terminals in a given production.
Through generating grammars with all possible combinations of these switches, we can create artificial languages of differing typological profiles.
This experimental paradigm allows us to conduct carefully controlled studies by varying only the typological parameter and make a causal claim.\looseness=-1

Using our method, we investigate inductive biases related to the head-directionality of several constructions.
We find that LSTM-based architectures show little bias towards any particular ordering, achieving similar average perplexities on all grammar variations tested.
This contradicts recent findings by \citet{ravfogel2019studying} who find LSTMs have a preference for SVO word order.
Conversely, we find that performance of transformer-based architectures varies significantly across our artificial languages; this is visualized in \cref{fig:kde}.
This indicates that some combinations of the switches result in languages with word orders that are harder for the transformer to model than others.
Our analysis suggests that neither the performance of the transformer-based architectures nor of the LSTM-based architectures reflects any known tendencies in attested natural languages, with the best performance being achieved on languages with the rarely-attested OVS sentence ordering.
Importantly, our method exposes that transformer-based language models and LSTM-based language models have vastly different inductive biases, a result that has not been clearly stated in the NLP literature.\looseness=-1

\section{Why Artificial Languages?}

\subsection{Previous Work}
Artificial languages have previously been used to investigate the ability of neural architectures with respect to specific phenomenon, such as their ability to acquire hierarchical generalizations \cite{mccoy-2018} and whether they can use systematic composition skills to make generalizations \cite{pmlr-v80-lake18a}.
\citet{bowman2015tree} also used artificial languages to investigate the ability of LSTMs to learn compositional structure, and compare their ability to that of tree-structured models.\looseness=-1

The work most closely related to ours is that of \citet{ravfogel2019studying}.
Taking methodological inspiration from \citet{wang2016galactic}, they create artificial versions of English with modified word order and case systems, including a version with object--verb agreement.
They use the task of predicting the number of the subject and object of a missing verb to examine language model performance across these variations.
They find that the models perform better on this task for the language with SVO word order. 
What they leave unchanged in their experiment, however, is the original English ordering within the constituents, e.g. the adjective--noun
ordering in a noun phrase.
However, constituent order correlates with ordering of other grammatical constituents typologically \citep{greenberg1963universals}, and this could lead to unwarranted preferences for the original English ordering.
Our work addresses this problem by using fully artificial languages rather than modifying English sentences.
This allows for our experiment to be more controlled by eliminating possible confounders.

Other work conducted on the topic of inductive biases of language models has tended to focus on correlational studies investigating the relationship between typological features extracted from the World Atlas of Language Structures  \citep[WALS;][]{wals}, which have only found negative results \citep{cotterell-etal-2018-languages, mielke-etal-2019-kind}.
Since this work looked exclusively at the features of attested natural languages, it is difficult to control for the multiple typological dimensions along which any two natural languages differ.
Further, given the large number of typological features exhibited among the world's languages, there are simply not enough attested languages to make strong correlational claims.
\citet{mielke-etal-2019-kind} ultimately 
concluded with a negative result; this negative result, in part, motivates our study.

\subsection{The Necessity of Artificial Languages}
We suggest that properly investigating the inductive biases of language models will likely require artificial languages. 
Choosing languages to investigate the inductive bias of a language model requires a trade-off between the experiment being realistic and being controlled.
Using attested natural languages gives us the most realistic representation of natural language and all its complexities, but this also reduces the level of control and makes it difficult to disentangle the various typological variables that differ between languages.
Indeed, this was the conclusion of \citet{mielke-etal-2019-kind}. 
Work such as \citet{ravfogel2019studying} finds some mid-point by using artificial languages which have been modified from English.
This means that the language is less natural and more controlled, but does not maximize either.\looseness=-1

In our experiments, we have chosen to maximize the level of control.
This means that our grammars are simple and do not necessarily cover all possible constructions that one would expect to see in a natural language.
However, our reward for this sacrifice is that we can precisely control and understand how two languages tested differ from one another.
We argue that this provides a good base for the exploration of inductive bias, as when differences are observed under these conditions we may now make a causal claim about their origin.
In future work, the base grammars could be changed and extended as much as necessary to test additional hypotheses.

\begin{figure}[t!]
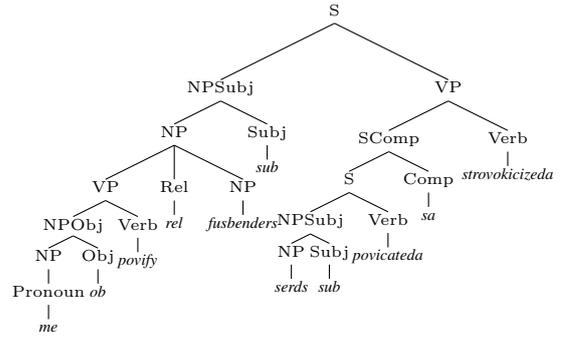
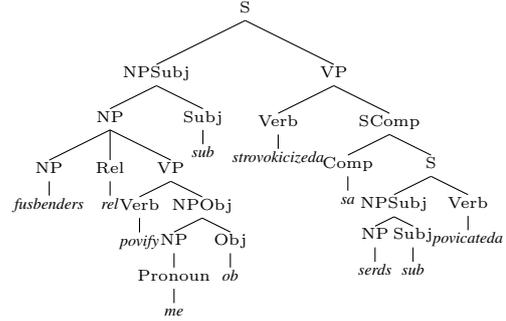
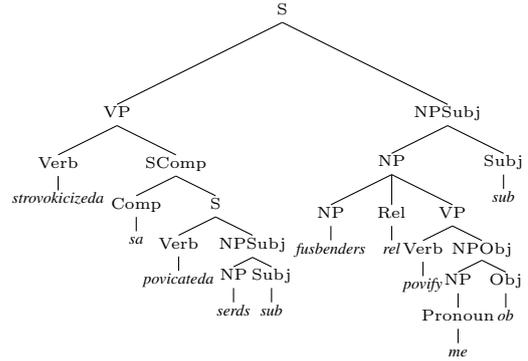

\centering
\begin{subfigure}{.5\textwidth}
\tiny

    \Tree 
    [.{$\Ss$} 
        [.{$\mathrm{NPSubj}$}
            [.{$\NP$}  
                [.{$\VP$}  
                    [.{$\mathrm{NPObj}$} 
                        [.{$\NP$} 
                            [.{$\mathrm{Pronoun}$} \textit{me} ] 
                        ] !\qsetw{0.6cm}
                        [.{$\Obj$} \textit{ob} ] !\qsetw{0.7cm}
                    ] !\qsetw{0.75cm}
                    [.{$\mathrm{Verb}$} \textit{povify} ] 
                ]  
                [.{$\Rel$} \textit{rel} ] !\qsetw{0.01cm}
                [.{$\NP$} \textit{fusbenders} ] !\qsetw{0.001cm}
            ] !\qsetw{1.5cm}
            [.{$\Subj$} \textit{sub} ] 
        ] !\qsetw{1cm}
        [.{$\VP$}  
            [.{$\mathrm{SComp}$} 
                [.{$\Ss$} 
                    [.{$\mathrm{NPSubj}$} 
                        [.{$\NP$}  \textit{serds} ] !\qsetw{0.5cm}
                        [.{$\Subj$}  \textit{sub} ] !\qsetw{0.5cm}
                    ] 
                    [.{$\mathrm{Verb}$} \textit{povicateda} ] !\qsetw{0.6cm}
                ] !\qsetw{1cm}
                [.{$\Comp$} \textit{sa} ] 
            ] !\qsetw{1.5cm}
            [.{$\mathrm{Verb}$} \textit{strovokicizeda} ]
        ]  
    ]
  \caption{Grammar 000000: me ob povify rel fusbenders sub serds\\ sub povicateda sa strovokicizeda .}
  \label{fig:000000}
\end{subfigure}
\begin{subfigure}{.5\textwidth}
\tiny
    \Tree 
    [.{$\Ss$}  
        [.{$\mathrm{NPSubj}$} 
            [.{$\NP$}  
                [.{$\NP$} \textit{fusbenders} ] !\qsetw{0.5cm}
                [.{$\Rel$} \textit{rel} ] !\qsetw{0.5cm}
                [.{$\VP$} 
                    [.{$\mathrm{Verb}$} \textit{povify} ] 
                    [.{$\mathrm{NPObj}$}  
                        [.{$\NP$}  
                            [.{$\mathrm{Pronoun}$} \textit{me} ] !\qsetw{0.1cm}
                        ] 
                        [.{$\Obj$} \textit{ob} ] !\qsetw{0.1cm}
                    ] !\qsetw{0.7cm}
                ] !\qsetw{1.1cm}
            ] !\qsetw{1.5cm}
            [.{$\Subj$} \textit{sub} ] 
        ] 
        [.{$\VP$} 
            [.{$\mathrm{Verb}$} \textit{strovokicizeda} ] 
            [.{$\mathrm{SComp}$} 
                [.{$\Comp$} \textit{sa} ] 
                [.{$\Ss$}  
                    [.{$\mathrm{NPSubj}$} 
                        [.{$\NP$} \textit{serds} ] !\qsetw{0.5cm}
                        [.{$\Subj$}  \textit{sub} ] !\qsetw{0.5cm}
                    ] !\qsetw{0.6cm}
                    [.{$\mathrm{Verb}$} \textit{povicateda} ] 
                ] !\qsetw{1.1cm}
            ] !\qsetw{1.3cm}
        ] !\qsetw{2.5cm}
    ]
  \caption{Grammar 011101: fusbenders rel povify me ob sub \\strovokicizeda sa serds sub povicateda .}
  \label{fig:011101}
\end{subfigure}
\begin{subfigure}{.5\textwidth}
\tiny

    \Tree 
    [.{$\Ss$}  
        [.{$\VP$}  
            [.{$\mathrm{Verb}$} \textit{strovokicizeda} ] !\qsetw{2cm}
            [.{$\mathrm{SComp}$}  
                [.{$\Comp$} \textit{sa} ] 
                [.{$\Ss$}   
                    [.{$\mathrm{Verb}$} \textit{povicateda} ] 
                    [.{$\mathrm{NPSubj}$} 
                        [.{$\NP$} \textit{serds} ] !\qsetw{0.5cm}
                        [.{$\Subj$}  \textit{sub} ] !\qsetw{0.5cm}
                    ] !\qsetw{0.6cm}
                ] !\qsetw{1cm}
            ] !\qsetw{1.1cm}
        ] !\qsetw{4.5cm}
        [.{$\mathrm{NPSubj}$} 
            [.{$\NP$}
                [.{$\NP$} \textit{fusbenders} ] !\qsetw{0.1cm}
                [.{$\Rel$} \textit{rel} ] !\qsetw{0.1cm}
                [.{$\VP$}  
                    [.{$\mathrm{Verb}$} \textit{povify} ] 
                    [.{$\mathrm{NPObj}$}  
                        [.{$\NP$} 
                            [.{$\mathrm{Pronoun}$} \textit{me} ] 
                        ] !\qsetw{0.4cm}
                        [.{$\Obj$} \textit{ob} ] 
                    ] !\qsetw{0.6cm}
                ] !\qsetw{1.5cm} 
            ] !\qsetw{2cm}
            [.{$\Subj$}  \textit{sub} ]
        ] !\qsetw{4.2cm}
    ]
  \caption{Grammar 111111: strovokicizeda sa povicateda serds \\ sub fusbenders rel povify me ob sub .}
  \label{fig:111111}
\end{subfigure}%
\caption{Trees showing the structure of parallel sentences across 3 of our artificial languages}
\label{fig:trees}
\end{figure}

\section{Constructing Controlled Languages}

\subsection{A Fully Controlled Experiment}%
A context-free grammar (CFG) is a quadruple ($\calN$, $\Ss$, 
$\Sigma$, $\calR$) where $\calN$ is a set of non-terminals, $\Ss \in \calN$ is a distinguished start non-terminal, $\Sigma$ is an alphabet and $\calR$ is a set of production rules.
An element $r \in \calR$ takes the form $N \rightarrow \valpha$ where $\valpha \in (\calN \cup \Sigma)^*$.
A CFG defines a subset of $\Sigma^*$. 

Probabilistic context-free grammars (PCFG) are a probabilistic generalization of CFGs. 
Rather than simply defining a subset of $\Sigma^*$, a PCFG gives us a probability distribution of $\Sigma^*$
where the structure of the grammar gives us the structural zeros of the distribution. %
Given a PCFG, we can take samples from it in order to generate sentences.

We set out to construct a set of PCFGs to expose the inductive bias of neural language models.
These grammars are parametrized by several ``switches'', which determine the ordering of constituents within the grammar.
The ``switches'' used are described in more detail in \cref{section:variation}.

We write an initial base PCFG in which productions are written to correspond with the ordering obtained when all switches are ``off''.\footnote{The choice of which permutation is ``on'' or ``off'' is arbitrary. In this case, ``off'' switches correspond to head-final orderings.}
In this base PCFG, the rules which are affected by the toggling of each switch are marked.
From this, sentences are sampled.
On generation, each production in these sentences is marked with the switch it is associated with.
We then work through every combination of switches, replicating this same set of generated sentences and reversing productions as required by the switches, to produce multiple parallel corpora, identical in their content up to a reordering of constituents.\looseness=-1

This experimental set-up allows us to ensure that sentences in the corpus for each of our artificial %
languages differ only in the configuration of the switches.
In this way we can be confident in attributing any differences in performance to a causal difference in these switches rather than any differences caused by confounders, e.g. content, style or complexity of the sentences.

\subsection{Our Context-Free Grammar}%

Now we describe the construction of the PCFG with which
we experiment in this work. %
Example sentences from several of our generated languages are shown in \cref{fig:trees}.
The base grammar and the scripts for sampling from it and generating corpora for all switch configurations will be released at \url{https://github.com/rycolab/artificial-languages}.

\paragraph{The Alphabet $\Sigma$.}
Open-class words were taken from a list of phonotactically plausible English pseudowords \cite{kharkwal2014taming}.
These pseudowords included verbs, nouns and adjectives.
We inflected the nouns manually for %
English plurality (adding \textit{s} or \textit{es}) depending on what English phonotactics requires.
We conjugated the verbs for present and past tense, again, using the rules of English.
Additional morphological markers that are not present in English, e.g. subject and object markers and an additional marker to denote a plural past tense verb form, were obtained by randomly sampling two-letters slices from the list of morphological plausible words.\footnote{This sampling occurred only once, and markers used were the same for all words.}
Pronouns and prepositions were also obtained in this fashion. 

\paragraph{The Non-Terminals $\calN$.}
Our grammar has a single distinguished start symbol $\Ss$.
It describes verb phrases ($\VP$), containing transitive and intransitive verbs, as well as verbs that take a sentential complement (complementizers are denoted $\Comp$).
Nouns are marked as being objects or subjects using a particle (denoted $\mathrm{Obj}$ or $\mathrm{Subj}$).
Verbs in our grammar have two tenses (past and present).
Noun phrases ($\NP$), including those modified by adjectives ($\Adj$), relative clauses (where relativizers are denoted $\Rel$) and prepositional phrases ($\PP$), are described in our grammar.

\paragraph{The Production Rules $\calR$.}
Our production rules $\calR$ cover several common productions seen in natural language.
We list the production rules which are subject to switching in our experiment in \cref{tab:switch_rules}.

\paragraph{Modeling Morphological Agreement.}%
Our grammar models a simple form of morphological agreement: verbs agree with their subjects in number (singular or plural).
This introduces an element of long-term dependencies into our languages -- if a language model is to correctly predict a verb form, it must carry information about the number of the subject.
In order to enforce this agreement in our grammar, non-terminals are subscripted with their number (where applicable).

\paragraph{Assigning Probabilities.}
Weights given to each production were chosen manually through experimentation.
Some principles for choosing weights for a grammar in this manner are described by \citet{eisner2008competitive}. %
An automated method of assigning weights could be explored in future work.

\begin{table}[]
\small
\begin{tabular}{lll}
\toprule
& \multicolumn{2}{l}{\textbf{Rule for each switch value}}                                                                \\\midrule
\textbf{Switch} & \textbf{0}                                                                                & \textbf{1}                                                                                \\\midrule
${\boldsymbol \Ss}$           & $\Ss\rightarrow\NP\, \VP$                                                                     & $\Ss \rightarrow \VP\, \NP$                                                                     \\
${\boldsymbol \VP}$          & $\VP \rightarrow \NP\, \VP$                                                                    & $\VP \rightarrow \VP\, \NP$                                                                    \\
${\boldsymbol \Comp}$        & $\Ss_\Comp \rightarrow \Ss\, \Comp$                                                              & $\Ss_\Comp \rightarrow \Comp\,\Ss$                                                           \\
${\boldsymbol \PP}$          & \begin{tabular}[c]{@{}l@{}}$\NP \rightarrow \PP\, \NP$\\ $\PP \rightarrow \NP\, \Prep$\end{tabular} & \begin{tabular}[c]{@{}l@{}}$\NP \rightarrow \NP\, \PP$\\$\PP \rightarrow \Prep\, \NP$\end{tabular} \\
${\boldsymbol \NP}$          & $\NP \rightarrow \Adj\, \NP$                                                                   & $\NP \rightarrow\NP\, \Adj$                                                                 \\
${\boldsymbol \Rel}$         & $\NP \rightarrow \VP\, \Rel\, \mathrm{Noun}$                                                              & $\NP \rightarrow \mathrm{Noun}\,\Rel\,\VP$                                                              \\ \bottomrule
\end{tabular}
\caption{Rules that are switchable in our grammar. Subscripts for tense and number agreement are not shown for simplicity.}
\label{tab:switch_rules}
\end{table}

\subsection{Controlled Typological Variation}\label{section:variation}%

\begin{CJK}{UTF8}{min}

\begin{table*}[]
\centering
\fontsize{8}{9.5}\selectfont
\setlength{\tabcolsep}{1pt}
\begin{tabular}{lclclcl}
\toprule
&      \multicolumn{2}{l}{\textbf{Japanese}}& \multicolumn{2}{l}{\textbf{English}}               & \multicolumn{2}{l}{\textbf{Spanish}}                                                           \\\midrule
\textbf{Switch}       & \textbf{Value} & \textbf{  Example} & \textbf{Value} & \textbf{  Example} & \textbf{Value} & \textbf{  Example}                  \\\midrule
${\boldsymbol \Ss}$      &  0 & {\color{red}猫}が{\color{orange}食べる}。    & 0      & {\color{red}The cat} {\color{orange}eats}.                  & 0 & {\color{red}El gato} {\color{orange}come}.                     \\
${\boldsymbol \VP}$    & 0 & 猫が{\color{red}ネズミ}を{\color{orange}食べる}。      & 1                         & The cat {\color{orange}eats} {\color{red}the mouse}.    & 1     & El gato {\color{orange}come} {\color{red}el ratón}.            \\
${\boldsymbol \Comp}$  & 0 & {\color{red}猫が食べる}{\color{orange}と}思う。       & 1                 & I think {\color{orange}that} {\color{red}the cat eats}.  & 1    & Pienso {\color{orange}que} {\color{red}el gato come}.\\
${\boldsymbol \PP}$  & 0   & {\color{cyan}テーブル}の{\color{orange}上}の{\color{red}猫}が食べる。                 & 1           & {\color{red}The cat} {\color{orange}on} {\color{cyan}the table} eats.  & 1    & {\color{red}El gato} {\color{orange}sobre} {\color{cyan}la mesa} come.       \\
${\boldsymbol \NP}$  & 0  & {\color{orange}小さな}{\color{red}猫}が食べる。             & 0                   & The {\color{orange}small} {\color{red}cat} eats.         & 1    & El {\color{red}gato} {\color{orange}pequeño} come.             \\
${\boldsymbol \Rel}$ & 0   & {\color{orange}ミルクを飲む}{\color{red}猫}が食べる。     & 1                        & {\color{red}The cat} that {\color{orange}drinks milk} eats.  & 1 & {\color{red}El gato} que {\color{orange}bebe leche} come.     \\\bottomrule
\end{tabular}
\caption{Demonstration of the orders of the switch constituents in Japanese, English and Spanish}
\label{tab:example_orders}
\end{table*}
\end{CJK}
Our end goal is to construct a grammar parameterized 
by a binary vector of $K$ switches.
We denote such a vector of switches $\vb \in \{0, 1\}^K$. %
Toggling an individual switch in the grammar reverses the order
of the right-hand sides of a set of production rules.
For example, the switch that we term the $\Ss$ switch reverses the order of the production $\Ss \rightarrow \NP\,\VP$ to create $\Ss \rightarrow \VP\,\NP$.\footnote{Details of all switches are shown in \cref{tab:switch_rules}.}
$2^K$ different grammars are possible from $K$ binary switches.
In the following paragraphs, we describe each of the switches
we consider in this work.

\paragraph{Position of subject in sentence ($\Ss$ Switch).}
This switch determines the order in which a subject and its verb phrase appear within a sentence.
If the switch has a value of $0$, the rule $\Ss \rightarrow \NP\,\VP$ is used, which is the order used in the vast majority of the world's languages, including SVO languages such as English and SOV languages such as Japanese.
If the switch has a value of $1$, the rule becomes $\Ss \rightarrow \VP\,\NP$.
This order is rare among attested natural languages, but can be seen in VOS languages such as Malagasy and OVS languages such as Hixkaryana. %

\paragraph{Position of verb in verb phrase ($\VP$ Switch).}
This switch determines whether a direct object precedes or follows its verb.
If the switch has a value of $0$, we use the head-final order, with the object preceding the verb.
This is seen in languages such as Japanese and Turkish.
If the switch has a value of $1$, the head-initial order is used, with the object following the verb.
This is seen in languages such as English and Chinese.
This switch, in combination with the $\Ss$ switch, determines the overall ordering of subject, object and verb within a sentence.
If the values of these switches are $(0,0)$, the language will have SOV word order, like Japanese and Turkish.
If they are $(1,1)$, the language will have VOS order, which is rare but can be seen in languages such as Malagasy.
SVO languages such as English correspond to $(0,1)$.
$(1,0)$ corresponds to OVS order, which is attested in only a very small number of human languages.

\paragraph{Position of complementizer in sentential complement ($\Comp$ switch).}
This switch determines whether a complementizer begins or ends a sentential complement.
If the switch has a value of $0$, the complementizer appears in head-final position, at the end of the complement.
This is the order seen in Japanese.
If the switch has a value of $1$, the complementizer appears in head-initial position, at the beginning of the complement.
This is the order seen in English.

\paragraph{Ordering of prepositional phrase ($\PP$ Switch).}%
This switch determines the ordering of a prepositional phrase.
If the switch has a value of $0$, the prepositional phrase precedes the noun it modifies, and the prepositional phrase ends with a preposition, in head-final order.
This order is seen in Japanese.  %
If the switch has a value of~$1$, the prepositional phrase follows the noun it modifies, and the preposition begins the prepositional phrase, in head-initial order.
This order is seen in English.\vspace{-5pt}

\paragraph{Position of adjective in noun phrase ($\NP$ Switch).}
This switch determines whether an adjective appears before or after the noun it modifies.
If the switch is $0$, the adjective precedes the noun (as in English and Japanese) and if it is $1$, the adjective follows the noun (as in Spanish and Irish).

\paragraph{Position of relative clause ($\Rel$ switch).}
This switch determines the position of a relative clause with respect to the noun it modifies.
If the switch has a value of $0$, a relative clause is followed by a relativizer and then the noun it modifies.
This order is seen in Japanese.
If the switch has a value of $1$, the noun being modified appears first, followed by a relativizer and the clause.
This order is seen in French and English.
\begin{figure*}[t!]
    \centering
    \includegraphics[scale=0.25]{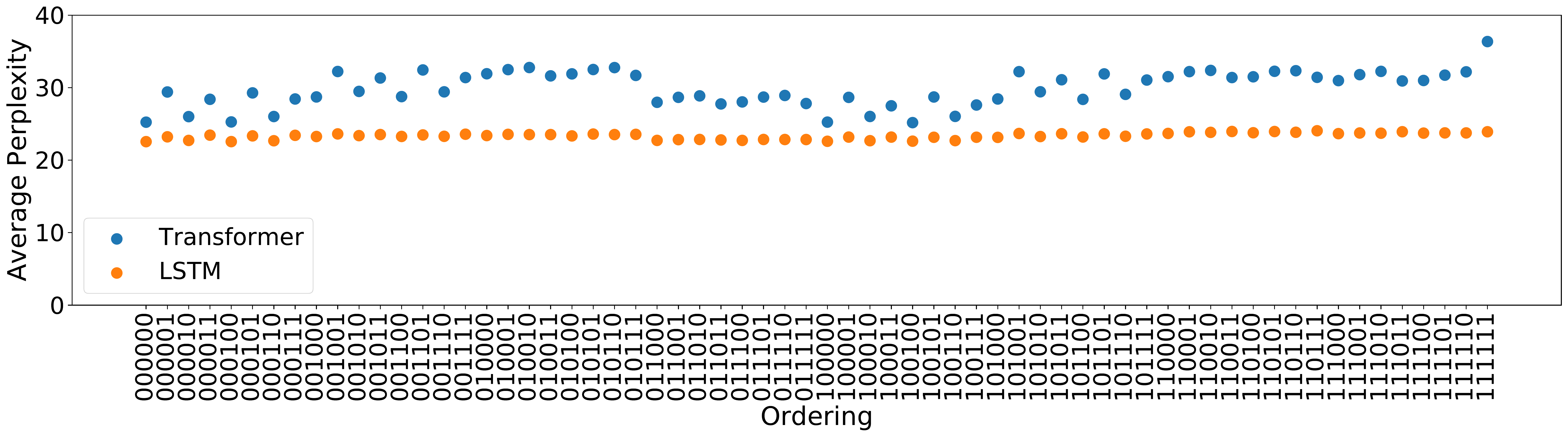}
    \caption{All scores achieved by LSTM- and transformer-based models}
    \label{fig:scatter}
\end{figure*}%

The unmarked word order of some attested languages can be approximately identified with particular switch vectors.\footnote{This is, of course, a simplification, since word order within a natural language can follow more complex rules, or allow for flexibility.} %
For example, standard English order corresponds approximately to $(0,1,1,1,0,1)$, Japanese to $(0,0,0,0,0,0)$ and Spanish to $(0,1,1,1,1,1)$.\footnote{From this point on, grammars will be referred to by their configuration of switches, sans brackets, e.g. Grammar 011101.} %
This is demonstrated in \cref{tab:example_orders}.
We note that our configurations cannot account for all possible word orders seen in attested languages (VSO languages are not represented, for example), but constitute a subset of possible orders.

\section{Experiments}
\paragraph{Architectures and Data.}
In order to compare inductive biases across architectures, two neural architectures were tested: transformers and LSTMs.
We used the implementation available as part of Fairseq \citep{ott2019fairseq}.
Our base grammar has $K=6$ switches, i.e. 6 binary choice points as described in \cref{section:variation}. This results in $2^6=64$ possible grammars.
For each of these grammars we generated 100,000 sentences, which were divided into 10 splits of 10,000.\footnote{10,000 sentences may sound like a relatively small number, but we note that our artificial languages are simple with small vocabularies, so we consider this number to be sufficient.}
The sentences generated for each grammar differed only in the designated choice points, i.e. in the ordering of their constituents.
This meant that each sentence appeared in an equivalent form in each grammar.
As such, for each sentence, we can compare the perplexity of the 64 variants of the sentence as calculated by language models trained on the corresponding grammars.
Each split of 10,000 sentences was divided into an 80--10--10 train--dev--test split.\footnote{Equivalent sentences across grammars were assured to be in the equivalent splits for each grammar, so train, dev and test sets across grammars contained the same sentences up to reordering of constituents.}

\paragraph{Procedure.}
We trained both a transformer-based and an LSTM-based language model on each train split and the models were evaluated on the test split.
This procedure resulted in 10 language models per architecture for each possible grammar, each of which was evaluated on 1,000 sentences in their respective test set.
The perplexity achieved on these test sets was averaged across the 10 splits, to give the average perplexity for that grammar. %
This approach helps to account for the variability between individual training runs.

\section{Results and Analysis}%
\subsection{Perplexity Evaluation}

\begin{figure*}
\centering
\begin{subfigure}{.5\textwidth}
  \centering
  \includegraphics[scale=0.2]{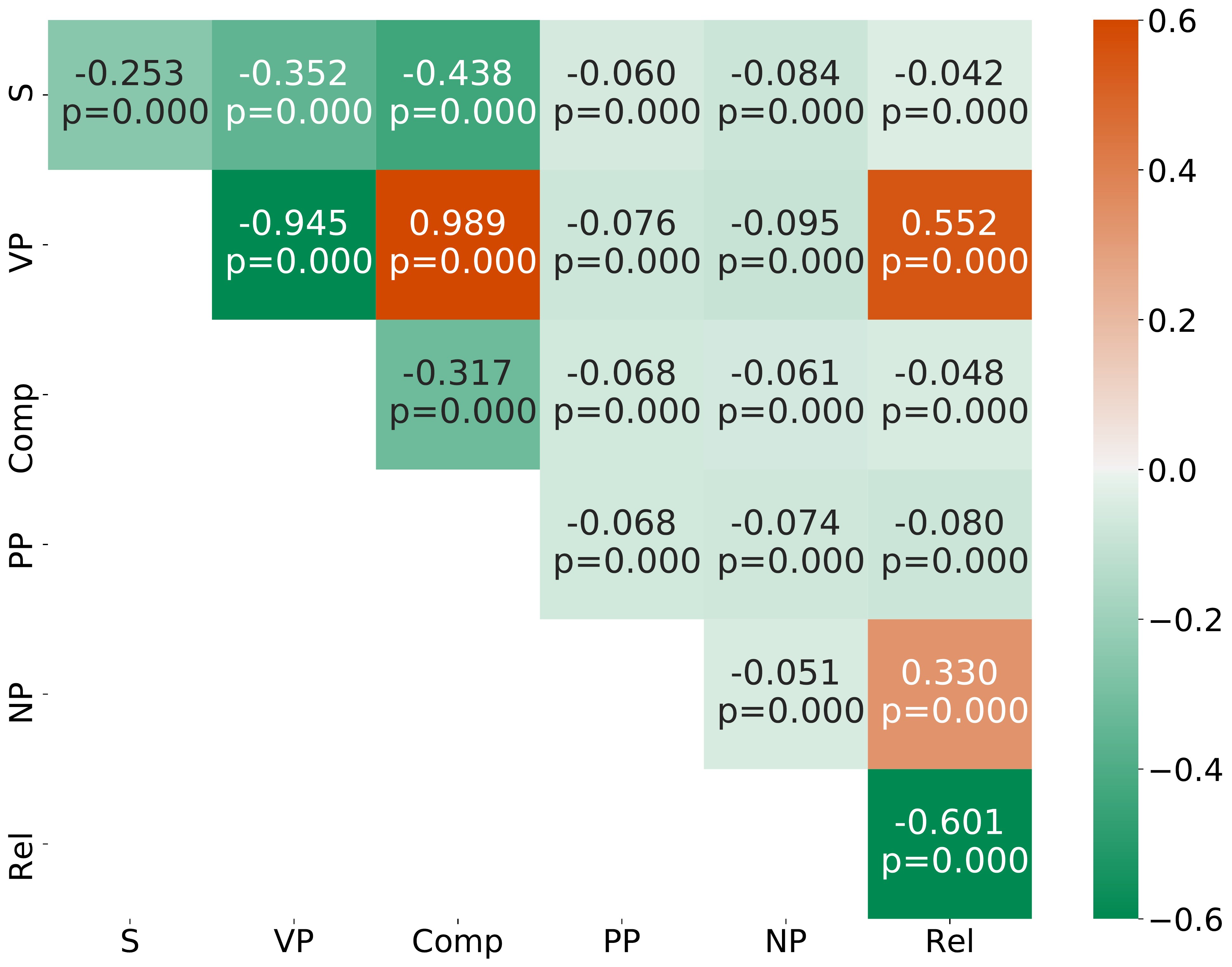}
  \caption{}
  \label{fig:heat_trans}
\end{subfigure}%
\begin{subfigure}{.5\textwidth}
  \centering
  \includegraphics[scale=0.2]{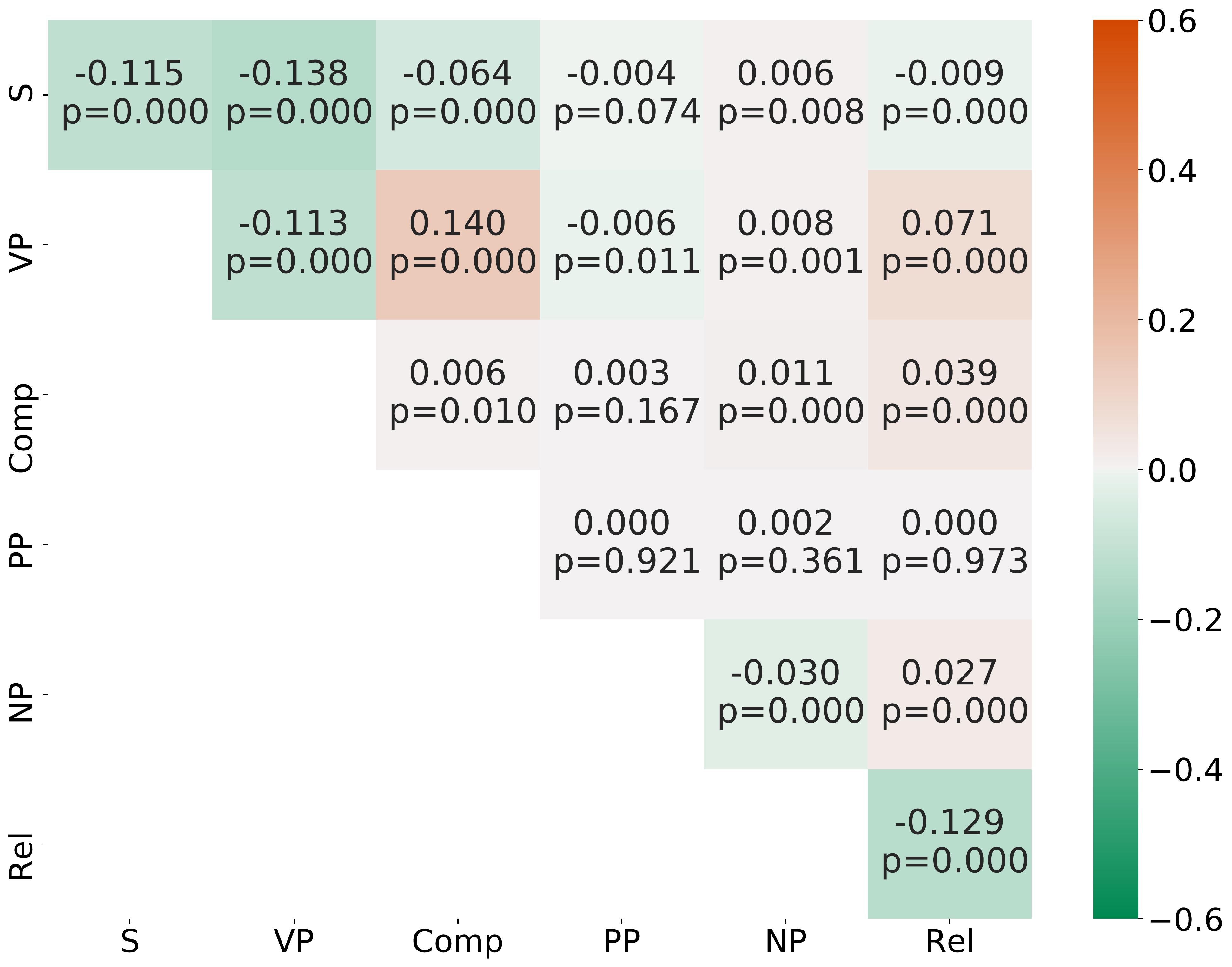}
  \caption{}
  \label{fig:heat_lstm}
\end{subfigure}
\caption{Heat maps showing the coefficients obtained for a mixed-effects model for perplexity as predicted by (a) transformers and (b) LSTMs.}
\label{fig:heat_maps}
\end{figure*}

The average perplexity on the test set was measured for each grammar.
This measures how well a language model explains
the held-out test set.
The lower the perplexity the better the language model fits the held-out data.
Average perplexity achieved across all grammars by the transformer- and LSTM-based models are shown in \cref{fig:scatter}.\footnote{Error bars are omitted, but across grammars the error on each measurement is generally between 0.25 and 0.5.}

\subsection{Mixed-Effects Modeling}%
We use a linear mixed-effects model to investigate the effects of each choice point in the grammar.
This allows us to model the effect of each switch in the grammar, and first-order interaction terms between them, on the perplexity of a sentence, while controlling for the fact that perplexities for parallel sentences across grammars are related (by using a random intersect per sentence grouping).
This model is explained in detail below.

Assume we have $N$ paired sentences 
from each of our $2^K$ grammars.
Let ${\boldsymbol L} \in \mathbb{R}_{\geq 0}^{N \times 2^K}$ be a non-negative real matrix of the perplexity obtained for every test sentence across every grammar.
Specifically, we have that $L_{nk}$ is the perplexity for the $n^{\text{th}}$ sentence
under the $k^{\text{th}}$ grammar.
Furthermore, let ${\boldsymbol S} \in \{0, 1\}^{2^K \times \left(\frac{K(K-1)}{2} + K\right)}$ be the binary matrix containing the configuration of switches and the $\frac{K(K-1)}{2}+K$ switch--switch interactions for each of the $2^K$ grammars in contrast coding \cite{wu2009mixed}. %
Thus, 
we have that the column vector ${\boldsymbol S}_{k\bullet}$ is a binary vector of length $\frac{K(K-1)}{2} + K$.
Let ${\boldsymbol \beta} \in \mathbb{R}^{\frac{K(K-1)}{2} + K}$ be a vector of real coefficients to be estimated describing the effect of each switch and their interactions. 
Let $u_n \sim \mathcal{N}(0, \sigma_{\mathrm{dif.}}^2)$ 
be a sentence-specific difficulty term
(a random effect) and let $\varepsilon \sim \mathcal{N}(0, \sigma^2)$ %
be a sentence--grammar-specific noise term.
Now, we model an individual perplexity $L_{nk}$, which corresponds to the $n^{\text{th}}$ sentence and the $k^{\text{th}}$ grammar, as follows:
\begin{equation} \label{eq:mem}
L_{nk} = {\boldsymbol S}_{k\bullet}\cdot{\boldsymbol \beta} + u_n + \varepsilon
\end{equation} %
Importantly, we draw one $u_n$ for each unique sentence. It is in this sense that $u_n$ acts as a term for modeling sentence difficulty. 
We may write \cref{eq:mem} as the following
\begin{equation}
    L_{nk} \sim \mathcal{N}({\boldsymbol S}_{k\bullet}\cdot{\boldsymbol \beta}, \sigma^2_{\mathrm{dif.}} + \sigma^2)
\end{equation}
which reveals that it is no more than a simple Gaussian model with tied parameters. 
We estimate ${\boldsymbol \beta}$, $\sigma^2_{\mathrm{dif.}}$ and $\sigma^2$ through maximum-likelihood estimation, which, in Gaussian models, is
equivalent to least-squares estimation.
A positive coefficient $\beta_j$ for a given switch means that models perform worse with head-initial ordering for that switch, while a negative coefficient means the opposite. \looseness=-1%
Since the fixed effects were input using contrast coding, the interaction terms in our model deal with the effects of two constituents \emph{sharing head-directionality}.
A positive coefficient for an interaction means that the models perform \emph{worse} when they share head directionality, and a negative coefficient means the opposite.
Head-directionality is commonly correlated between sentence constituents in attested natural languages, so if the biases of these architectures reflected human languages, we would expect most interaction terms to be negative. %
The coefficients obtained for the transformers are shown in \cref{fig:heat_trans}.
Those for the LSTMs are shown in \cref{fig:heat_lstm}. %

\section{Discussion}
\paragraph{Differences Between Architectures.}
It is clear from \cref{fig:scatter} that the transformer- and LSTM-based models do not show the same inductive biases with respect to the switches we investigated.
Across all possible configurations of the switches, LSTMs achieve very similar average perplexities, suggesting that they have little preference for any particular set of constituent orderings.
In contrast, the average perplexities achieved by the transformers vary considerably between grammars.
This demonstrates clearly that the two models exhibit distinctly different preferences with regard to orderings of words within in a sentence.
Further, the clear contrast between the coefficients obtained by the mixed-effects models for transformers and LSTMs (shown in \cref{fig:heat_trans} and \cref{fig:heat_lstm}, respectively) demonstrates a stark contrast between the two models. %
None of the switches investigated, or their first-order interactions, appear to have a substantial effect on the scores obtained in the case of the LSTM-based models, whereas the transformer-based models are clearly affected to a much greater degree by the configuration of these switches.
Given that these two architectures are both commonly used for similar tasks, such a difference in their inductive biases is noteworthy.

\paragraph{Correlated Switches.}\jennifer{has been changed and needs checking}
\Cref{fig:heat_trans} shows the coefficients obtained by the mixed-effects model employed to investigate the effects of the switches on performance for the transformer-based models.
The diagonal values (for single switches) are all negative coefficients, which indicates that the model performance is better when these have head-final ordering.
Off-diagonal values are the coefficients obtained for the interaction terms between two switches.
A positive value here indicates that when these two switches have the same value (either both head-initial or both head-final), the performance of the model is worse.
A negative value means that when the two switches have the same value, the performance is better.
Most of the off-diagonal elements have small values, with a few exceptions.
The coefficients of the cross terms between the $\Ss$ and $\VP$ switches and the $\Ss$ and $\Comp$ switches are larger negative values, which indicates that when these constituents share their head-directionality the performance of the transformer-based models is better.
The coefficients of the cross terms between the $\VP$ and $\Comp$, $\VP$ and $\Rel$ and $\NP$ and $\Rel$ switches are larger postive values, indicating that the transformers perform worse when these constituents share head-directionality.
Generally, attested natural languages tend to exhibit a tendency towards one head-directionality, but the transformer does not seem to have inductive biases that reflect this.
The corresponding coefficients for the LSTM-based models, shown in \cref{fig:heat_lstm}, are all small, further demonstrating that the LSTMs are largely agnostic to word ordering.

\begin{figure}
    \centering
    \includegraphics[scale=0.2]{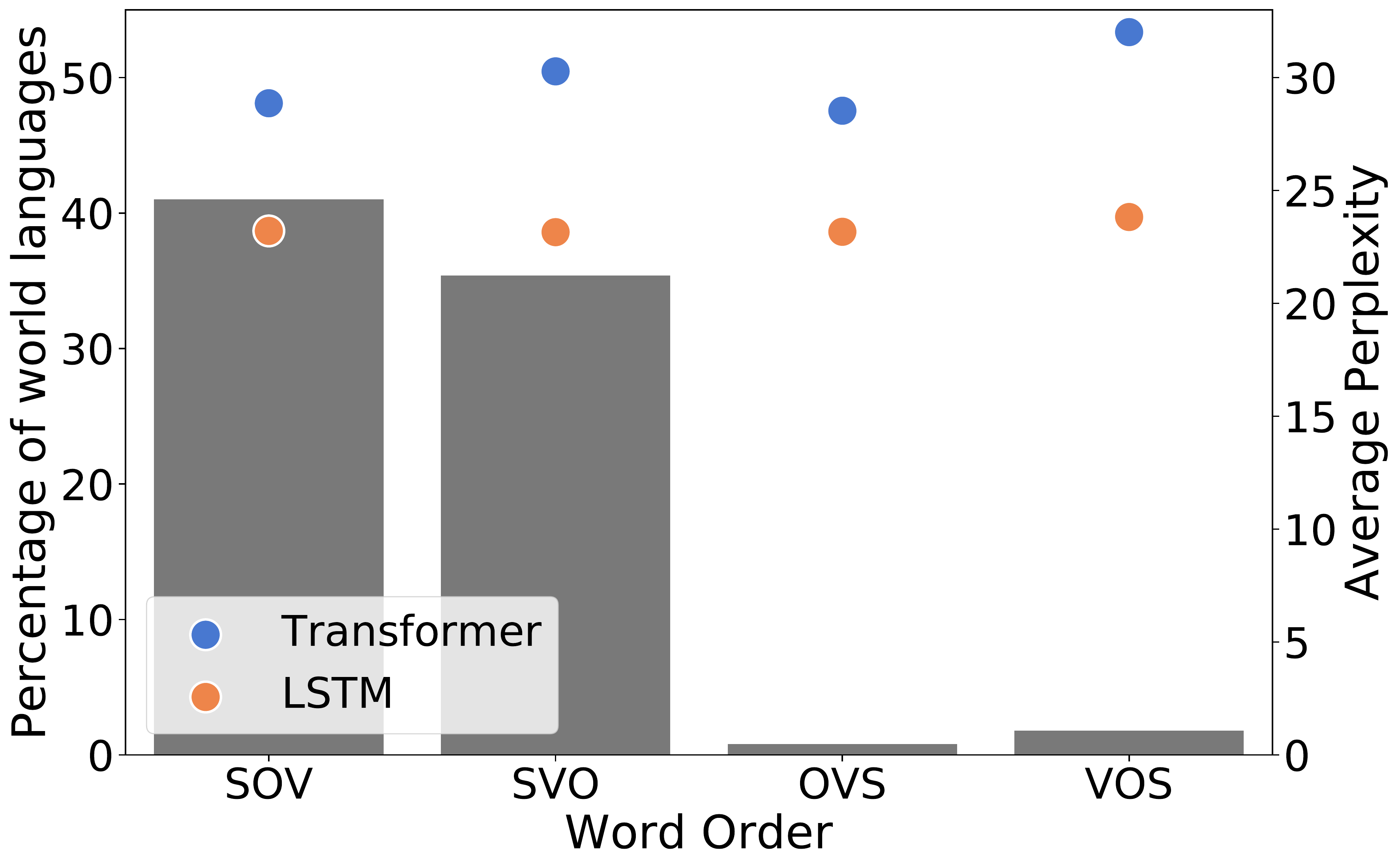}
    \caption{The prevalence of word orders across languages \cite{wals-81}, plotted with the average perplexities achieved on each of these groups of grammars by transformer- and LSTM-based models}
    \label{fig:orders}
\end{figure}

\paragraph{Tendencies in Attested Natural Languages.}
We wish to consider the question of whether the biases of these models are in any way reflective of word order tendencies that we see across attested natural languages.
All word orders are not equally common among natural languages, and it is interesting to consider whether the word orders that these models are able to model more successfully are those which are more commonly seen in natural language.
Some have speculated that the skew of word orders in human languages could possibly be reflective of human cognitive biases \citep{culbertson2012learning, culbertson2019world}, so it would be interesting to see to what extent the inductive biases of these models reflects this skew.
Since LSTMs appear to show no preference for any word order over the others, they are clearly not reflective of attested tendencies in word order.
To attempt to answer this question for the transformers, we begin by comparing the performance of the models on subsets of grammars with the prevalence of similar languages among humans.
In \cref{fig:orders}, the grammars are grouped by how they order the verb, object and subject of a sentence, and the average perplexities achieved by the language models on each of these groups is shown. %
On the same figure, we display the estimated prevalence of these orderings among the world's languages \cite{wals-81}.
It is clear that these two things are not correlated, with the transformer performing similarly on SOV languages, the most common among the world's languages, and OVS languages, which are rarely attested. %
This shows that the bias exhibited by transformers does not reflect tendencies among attested languages.
A further indication of this is the lack of a strong preference for switches sharing head-directionality as shown in \cref{fig:heat_trans}.
In human languages, the headedness of constituents is often correlated \cite{greenberg1963universals}.
We would expect to see this through negative coefficients for interaction terms in the mixed-effects model for constituents whose orders commonly correlate.
However, we do not observe this for all correlations.
For example, we would expect the $\PP$ switch to show a strong preference for shared head-directionality with other switches, which we do not observe.

\section{Conclusion}
We propose a novel methodology for the investigation of the inductive bias of language models using the technique of creating carefully controlled artificial languages.
This approach allows for the elimination of differences in corpora between languages and means that typological variation between languages can be restricted exclusively to the typological features being investigated.
We use this methodology to investigate the inductive bias of two neural architectures which are commonly used for this task: LSTMs and transformers.
We found that these two models have starkly different inductive biases with respect to word order, with the LSTM showing little variation in performance across word order, while the performance of the transformer varied significantly across artificial languages.\looseness=-1

\section*{Acknowledgements}
We thank Simone Teufel for providing feedback on an early draft.

\section*{Ethical Considerations}
The authors foresee no ethical concerns with the research presented in this paper. 

\bibliographystyle{acl_natbib}
\bibliography{anthology,acl2021}

\end{document}